\documentclass[%
 aip,
rsi,%
 amsmath,amssymb,
preprint,%
]{revtex4-1}

\usepackage{graphicx}
\usepackage{dcolumn}
\usepackage{bm}
\usepackage{xcolor}
\usepackage{mathtools}

\newcommand{\CDYrevise}[1]{\textcolor{black}{#1}}
\usepackage[normalem]{ulem}

\begin{document}


\title{Deep learning delay coordinate dynamics for chaotic attractors from partial observable data}

\author{Charles D. Young}

\affiliation{
Department of Chemical and Biological Engineering, University of Wisconsin-Madison, Madison, Wisconsin 53706, USA
}%

\author{Michael D. Graham}
 \email{mdgraham@wisc.edu}
\affiliation{ 
Department of Chemical and Biological Engineering, University of Wisconsin-Madison, Madison, Wisconsin 53706, USA
}

\date{\today}

\begin{abstract}
A common problem in time series analysis is to predict dynamics with only scalar or partial observations of the underlying dynamical system. For data on a smooth compact manifold, Takens theorem proves a time delayed embedding of the partial state is diffeomorphic to the attractor, although for chaotic and highly nonlinear systems learning these delay coordinate mappings is challenging. We utilize deep artificial neural networks (ANNs) to learn discrete discrete time maps and continuous time flows of the partial state. Given training data for the full state, we also learn a reconstruction map. Thus, predictions of a time series can be made from the current state and several previous observations with embedding parameters determined from time series analysis. \CDYrevise{The state space for time evolution is of comparable dimension to reduced order manifold models. These are advantages over recurrent neural network models, which require a high dimensional internal state or additional memory terms and hyperparameters. We demonstrate the capacity of deep ANNs to predict chaotic behavior from a scalar observation on a manifold of dimension three via the Lorenz system. We also consider multivariate observations on the Kuramoto-Sivashinsky equation, where the observation dimension required for accurately reproducing dynamics increases with the manifold dimension via the spatial extent of the system.}.
\end{abstract}

\maketitle

\section{\label{sec:Intro}Introduction}

Many applications require the prediction of a time series with short term tracking and long term statistical accuracy from only observable data, such as modeling turbulent flows, \cite{wang2021state} weather, \cite{elsner1992nonlinear} rainfall, \cite{dodov2005incorporating}  protein configurations, \cite{topel2020reconstruction} and the stock market. \cite{chandra2021bayesian} The system is often governed by underlying differential equations on a smooth compact manifold of dimension $d_\mathcal{M}$. For an observation of the system with dimension $d_o$ in ambient Euclidean space, $u(t) \in \mathbb{R}^{d_o}$, Whitney's theorem proves that there is a diffeomorphic mapping to the manifold coordinates $h(t) \in \mathbb{R}^{d_\mathcal{M}}$ when the observation dimension satisfies $d_o > 2d_\mathcal{M}$. \cite{guillemin2010differential, sauer1991embedology} In this case, time prediction of the state can be performed from only the current observation, as demonstrated by data-driven approaches such as sparse regression, \cite{brunton2016discovering} and reduced order modeling. \cite{linot2020deep, linot2022data} In other approaches, the full state is encoded with the history of the system to improve short-time predictions. \cite{jaeger2004harnessing, vlachas2018data, pathak2018model} An advantage of reduced order modeling is the ability the ability to perform time evolution at low computational expense, which is essential in control applications. \cite{zeng2022data} In particular, autoencoders discover a latent space $h = \chi(u;\theta) \in \mathbb{R}^{d_\mathcal{M}}$ which approximates the minimum dimension manifold with trainable parameters $\theta$. \cite{linot2020deep, linot2022data}

For a partial observable $u_p(t) \in \mathbb{R}^{d_p}$ of dimension $d_p < 2d_\mathcal{M}$, the information contained in the current observation is insufficient to reconstruct the manifold. An alternative approach is to embed the state and its $m-1$ time delays $u_d(t) = [u_p(t),u_p(t-\tau),...,u_p(t-(m-1)\tau)] \in \mathbb{R}^{m \times d_p}$. Takens' theorem proves that for an embedding dimension $m > 2d_\mathcal{M}$ and nearly any choice of delay spacing $\tau$ there exists a diffeomorphic delay coordinate map to the manifold. \cite{takens1981detecting, sauer1991embedology} Takens' theorem was originally formulated for scalar observations $d_p = 1$, but generalizations for vector observations have been developed. \cite{deyle2011generalized} While Takens' theorem proves the existence of delay coordinate maps, it does not offer any guidance in determining these functions. In this work, we use neural networks (NNs) to learn these delay coordinate dynamics. We continue the idea of a minimal data-driven model for chaotic dynamics, \cite{linot2020deep, linot2022data} where the delay coordinate embedding $u_d$ serves as the reduced order model in the absence of full state data.

Before learning delay coordinate dynamics, embedding parameters must be chosen. Progress in time series analysis has provided techniques for generating optimal embeddings to reconstruct the manifold. \cite{kantz2004nonlinear, bradley2015nonlinear} The embedding dimension is generally estimated by false nearest neighbor (FNN) methods \cite{kennel1992determining, cao1997practical} and the delay spacing by the mutual information \cite{fraser1986independent} or correlation integral. \cite{kim1999nonlinear} Notably, these methods have primarily been applied to scalar observations of low-dimensional chaotic attractors, and the choice of $m$ and $\tau$ are made nearly independently. Modern approaches which can account for multivariate observations and which aim to improve reconstruction of the true attractor's topology have been developed,  \cite{garcia2005multivariate, nichkawde2013optimal, mccullough2015time, xu2019twisty, kramer2021unified} but there has been limited testing of these methods on chaotic attractors of dimension $d_\mathcal{M} > 3$. Time series analysis methods have also been built into data-driven methods for automatic generation of a latent space for time evolution, where the delay coordinate embedding is encoded to mask redundant time delays and promote orthogonality. \cite{jiang2017state, gilpin2020deep, ouala2020learning, wang2021reconstructing}

 With a suitable delay coordinate embedding, the mappings for time prediction and reconstruction can be approximated. Advancements in machine learning have motivated many data-driven approaches for time prediction of partial observables including random feature maps and data assimilation, \cite{gottwald2021combining, brajard2020combining} sparse regression, \cite{bakarji2022discovering} augmented latent space embeddings, \cite{ouala2020learning, ouala2022bounded}  closure modeling, \cite{jiang2020modeling, harlim2021machine} and neural ordinary differential equations. \cite{wang2021reconstructing} Delay coordinate embeddings have also been used to model nonlinear dynamics by a linear system using Koopman theory. \cite{brunton2017chaos, pan2020structure} Many of these approaches explicitly construct delay coordinate embeddings. \cite{gottwald2021combining, bakarji2022discovering, wang2021reconstructing} Others invoke Takens' theorem implicitly by use of recurrent neural networks (RNNs), which contain a memory term for embedding the state history. \cite{harlim2021machine} Most methods test the ability to predict chaotic dynamics on the Lorenz-63 attractor, with short time tracking for 5-10 Lyapunov times. More recently, some methods have been applied to the Lorenz-96 attractor \cite{jiang2020modeling, brajard2020combining} and the Kuramoto-Sivashinsky equation (KSE). \cite{harlim2021machine} These approaches use RNNs, which generate non-Markovian dynamical models. Therefore, they do not generally represent a reduced order model of the state. The state history must be parameterized into the architecture's memory, requiring memory hyperparameters in the case of LSTMs \cite{harlim2021machine} and reservoirs with a high internal dimension in the case of echo state networks. \cite{lu2017reservoir}

In addition to forecasting, data-driven methods often seek to perform reconstruction of the true attractor as a supervised learning process. \cite{brajard2020combining, bakarji2022discovering, harlim2021machine, lu2017reservoir} In particular, reservoir computers and closure models have successfully reconstructed the KS attractor. \cite{lu2017reservoir} However, reservoir computers are non-Markovian and require a high dimensional internal state compared to a diffeomorphic embedding in $2d_\mathcal{M}$ delay coordinates. If the full state is unavailable for training, autoencoders have been used for unsupervised reconstruction of a latent attractor. \cite{jiang2017state, gilpin2020deep, ouala2020learning, wang2021reconstructing} 

We propose a method using deep neural networks to learn delay coordinate maps from partial observable data. We perform supervised learning of a discrete time map, a continuous time flow (ordinary differential equation representation), and a reconstruction map for multivariate partial observations of chaotic dynamics. Compared to other approaches, we demonstrate the scaling of our method to higher dimensional chaotic systems via the KSE for $L = 22, d_\mathcal{M} = 8$ and $L = 44, d_\mathcal{M} = 18$. \cite{yang2009hyperbolicity, ding2016estimating, linot2020deep, linot2022data} Additionally, our model is Markovian and requires few hyperparameters. The only required inputs are the embedding dimension $m$ and the delay spacing $\tau$, both of which can in principle be determined before training the model.  In practice we find some empirical testing to be required for selecting emebedding parameters, although our results are insensitive to these choices. Within an appropriate range of $m$ and $\tau$, the autocorrelation function of the state observation and probability distribution function of state variables are quantitatively consistent. Short-time tracking exhibits some sensitivity to choice of embedding parameters because these metrics correspond closely to the training loss, which is easier to minimize for an optimal embedding. The neural network depth and width are small compared to reservoir networks, which require $10^3$ nodes to encode the state history for chaotic systems such as the KSE. \cite{pathak2018model, lu2017reservoir}

\begin{figure*}
    \centering
    \includegraphics[width=1.0\linewidth]{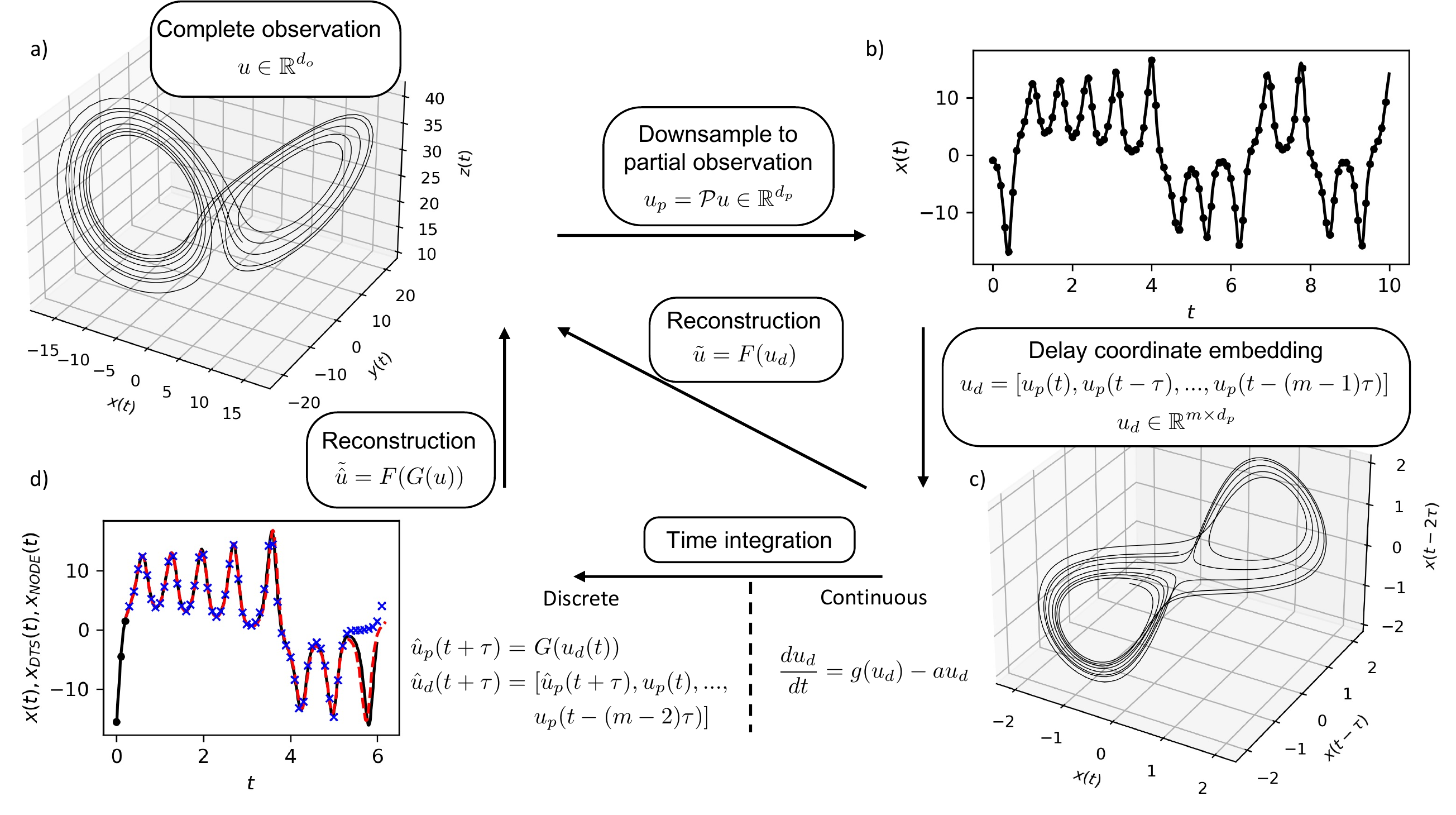}
    \caption{Schematic of learning delay coordinate dynamics and reconstruction a) Start with a complete observation of the attractor, here considering the Lorenz sytem b) Downsample to a partial observation, here taking $u_p = x$. Black points refer to the data at the delay coordinate spacing $\tau$ c) Construct a delay coordinate embedding. Here we select $m = 3$ delays and a delay spacing $\tau = 0.1$ d) Learn the discrete and continuous time dynamics of the embedding. Black points refer to the delay coordinate embedding initial condition, the black solid line to the data, blue crosses to the discrete time prediction, and the red dashed line to the NODE prediction. The true or predicted delay coordinate embedding can also be reconstructed to the full state given training data.}
    \label{fig:recon}
\end{figure*}

\section{\label{sec:Methods}Methodology}

We consider a full state space observation $u(t) \in \mathbb{R}^{d_o}$ from numerical simulations or, in principle, experimental data. We project to a lower dimension $d_p < d_o$ to generate a partial observation $u_p = \mathcal{P}u \in \mathbb{R}^{d_p}$. Here the projection operator simply filters out entries from the observation vector, but more general projections are compatible with our formulation. We construct a multivariate delay coordinate embedding of the partial observable and seek to learn the discrete time map and reconstruction map proposed by Takens theorem using deep neural networks (NNs). The data is available at a sampling interval $\Delta t$. We estimate the embedding dimension $m$ using false nearest neighbors \cite{kennel1992determining} and the delay spacing $\tau = n \Delta t$ using the first minimum of the mutual information, where $n$ is the number of samples between delay coordinates. \cite{fraser1986independent} The embedding is then defined as $u_d(t) = [u_p(t),u_p(t-\tau),...,u_p(t-(m-1)\tau)] \in \mathbb{R}^{m \times d_p}$.

\subsection{\label{sec:timestep}Discrete Time Evolution}
To advance the partial observable in time, we first consider a discrete time step (DTS) map:
\begin{equation}
\begin{split}
	\hat{u}_p(t+\tau) & = G(u_d(t);\theta_G), \quad G: \mathbb{R}^{m \times d_p} \rightarrow \mathbb{R}^{d_p} \\
	\hat{u}_d(t+\tau) & = [\hat{u}_p(t+\tau),u_p(t),...,u_p(t - (m-2)\tau)]
\end{split}
\end{equation}
where $\theta_G$ are NN parameters. The time step is equal to the delay time such that the delay coordinate vector can be iteratively forecasted. We approximate this function by a dense feed-forward neural network. Architecture details are given in Table \ref{table:arch}. NN weights are trained using stochastic gradient descent as implemented in Keras \cite{chollet2015keras} to minimize the loss:
\begin{equation}
	\mathcal{L}_G = \langle ||u_p(t+\tau) - \hat{u}_p(t + \tau)||^2_2 \rangle
\end{equation}
where $\hat{u}_p$ is the NN output.

 To generate long NN predicted trajectories, an initial delay coordinate embedding $u_d(0)$ is first integrated forward as $\hat{u}_p(\tau) = G(u_d(0);\theta_G)$. The new partial observation is used to update the delay coordinate embedding to $\hat{u}_d(\tau) = (\hat{u}_p(\tau),u_p(0),...,u_p(t-(m-2)\tau)$. The next time step is then calculated as $\hat{u}_p(2 \tau) = G(\hat{u}_d(\tau);\theta_G)$ and so on iteratively such that after $m$ steps the delay coordinate embedding contains only predicted values. An advantage of this approach is that only the current state and its $m-1$ delays are required to make predictions, in comparison to reservoir networks, which require a warmup period before predictions can be made. \cite{pathak2018model, lu2017reservoir}  However, intermediate time scales between the delay spacing are not accessible without interpolation.

\subsection{\label{sec:NODEs}Continuous Time Evolution}
We also consider the continuous time evolution of the delay coordinate embedding:
\begin{equation}
\frac{du_d}{dt} = g(u_d; \theta_g) - au_d
\end{equation}
where $g(u_d;\theta_g)$ is a neural network with parameters $\theta_g$, trained as described below, and the term $-a u_d$ has a stabilizing effect, keeping solutions from blowing up for appropriately chosen $a$. \cite{linot2022data, linot2022stabilized}  No generality is lost when including this term; the combination $g-a u_d$ is learned from the data. Unless otherwise noted, here $a = 10^{-3}$.
 Other than the damping term, our approach is similar to that of Wang and Guet. \cite{wang2021reconstructing} We approximate the delay coordinate dynamics $g$ by a neural network, or neural ordinary differential equation (NODE), which we use to integrate the state in time:  
\begin{equation}
\breve{u}_d(t + N \Delta t) = u_d(t) + \int_{t}^{t + N \Delta t} (g(u_d(t);\theta_g) - a u_d(t))dt,
\end{equation}
and the NN for $g$ is trained with the multi-step loss over $N$ steps of size $\Delta t$:
\begin{equation}
	\mathcal{L}_g = \left \langle \sum_{i=1}^{N} ||u_d(t + i \Delta t) - \breve{u}_d(t + i \Delta t) ||^2_2 \right \rangle.
\end{equation}
The gradient of the loss is calculated by backpropagating through the solver with automatic differentiation. \cite{chen2018neural}  The ODE solver uses the 5th order Dormand-Prince-Shapmine method implemented in torchdiffeq. \cite{chen2018neural}
\CDYrevise{In contrast to the discrete time case, where the model must be trained on a fixed time step and cannot resolve intermediate time scales, the neural ODE of the continuous time model can be trained and deployed for prediction on any time interval. Thus, the DTS model is trained on the time scale of interest, here the delay spacing $\tau$. The NODE is trained using the data sampling interval $\Delta t$ because previous work has shown smaller time steps improve training. \cite{linot2022data} However, the same study showed discrete and continuous time predictions were consistent for a data spacing $\Delta t < 0.5 \tau_L$, which is the case for all models trained in this work, so it is unlikely data spacing will significantly impact model performance.}  The partial observable can then be evaluated at times between delay spacings using the same solver used to determine $g$. For comparison to the DTS models, we report a NODE loss calculated after training using only the leading delay coordinate coordinate of the embedding as in the DTS loss:
\begin{equation}
	L_g ^\prime = \langle || u_p(t+\tau) - \breve{u}_{p}(t + \tau) ||^2_2 \rangle
\end{equation}

\subsection{\label{sec:recon}Reconstruction}
We also reconstruct the full observation from our numerical simulation data by delay coordinate mappings. We take a supervised learning approach on the assumption that training data is available. The reconstruction map from delay coordinates is defined as:
\begin{equation}
	\tilde{u}(t) = F(u_d(t);\theta_F), \quad F: \mathbb{R}^{m \times d_p} \rightarrow \mathbb{R}^{d_o}
\end{equation}
The mapping is again approximated by dense feed-forward NNs with weights $\theta_F$ as detailed in Table \ref{table:arch}. The reconstruction loss is:
\begin{equation}
	\mathcal{L}_F = \langle ||u(t) - \tilde{u}(t)||^2_2 \rangle
\end{equation}
where $\tilde{u}$ is the NN output. The reconstruction training is performed independently of the time integration training, separating the error associated with the two functions. A visual example of the reconstruction process is shown in Fig. \ref{fig:recon} for the diffeomorphism between the Lorenz attractor embedding of the partial state $u_p = x, m = 3, \tau = 0.1$ and the true attractor.

While the reconstruction training is performed only on true data, we apply the function to partial states predicted from the discrete time map. For investigating long time dynamics we will refer to reconstructions of NN predicted partial states as $\tilde{\hat{u}}(t) = F(\hat{u}_d(t);\theta_F)$, where predictions $\hat{u}_d(t)$ at long times $t$ are generated as described above.

\begin{table*}
\setlength{\tabcolsep}{12pt}
\centering
\begin{tabular}{c c c c}
\hline \hline
System & Function & Shape & Activation \\
\hline
Lorenz & $G$ & $m:200:200:1$ & ReLU:ReLU:linear \\
& $F$ & $m:200:200:3$ & ReLU:ReLU:linear \\
& $g$ & $m:200:200:200:m$ & ReLU:ReLU:ReLUlinear \\
KS $L=22$ & $G$ & $m d_p:256:256:256:256:d_p$ & ReLU:ReLU:ReLU:ReLU:linear \\
& $F$ & $m d_p:256:256:256:256:64$ & ReLU:ReLU:ReLU:ReLU:linear \\
& $g$ & $m d_p:256:256:256:256:m d_p$ & ReLU:ReLU:ReLUlinear \\
KS $L=44$ & $G$ & $m d_p:512:512:512:512:d_p$ & ReLU:ReLU:ReLU:ReLU:linear \\
& $F$ & $m d_p:512:512:512:512:64$ & ReLU:ReLU:ReLU:ReLU:linear \\
& $g$ & $m d_p:512:512:512:512:m d_p$ & ReLU:ReLU:ReLUlinear \\
\end{tabular}
\caption{NN architectures}
\label{table:arch}
\end{table*}

\section{\label{sec:Results}Results}

We apply our method to two common chaotic attractors to demonstrate the short term tracking and reproduction of long time statistics in the delay coordinate embedding space, as well as reconstruction of the long time statistics to the true attractor.

\subsection{Lorenz System}

We consider the Lorenz attractor, \cite{lorenz1963deterministic}
\begin{equation}
\begin{split}
	\frac{dx}{dt} & = \sigma(y - x) \\
	\frac{dy}{dt} & = x(\rho - z) - y \\
	\frac{dz}{dt} & = xy - \beta z
\end{split}
\end{equation}
where $\sigma = 10$, $\beta = 8/3$, and $\rho = 28$. The Lyapunov time using these parameters is $\tau_L \approx 1$, and the fractal dimension estimated by the correlation integral is is $d_A \approx 2.06$. \cite{grassberger1983characterization} The training data is generated using a Runge-Kutta 4-5 integrator in SciPy with a sampling time $\Delta t = 0.1$. The first $10^4$ data points are discarded as transients, and the next $5 \times 10^5$ data points are used for training with an 80/20 training/test split. The discrete time and reconstruction maps are trained in Keras \cite{chollet2015keras} using an Adam optimizer for 1000 epochs with an initial learning rate of 0.001, which is decreased by a factor of 0.5 every 100 epochs. The NODE models are trained using torchdiffeq, \cite{chen2018neural} with a batch size of 100 for $50,000$ epochs. The initial learning rate is 0.001, and it is decreased by a factor of 0.5 every 10,000 epochs. The number of time steps forecasted during training is $N = 2$.

\begin{figure*}
    \centering
    \includegraphics[width=1.0\linewidth]{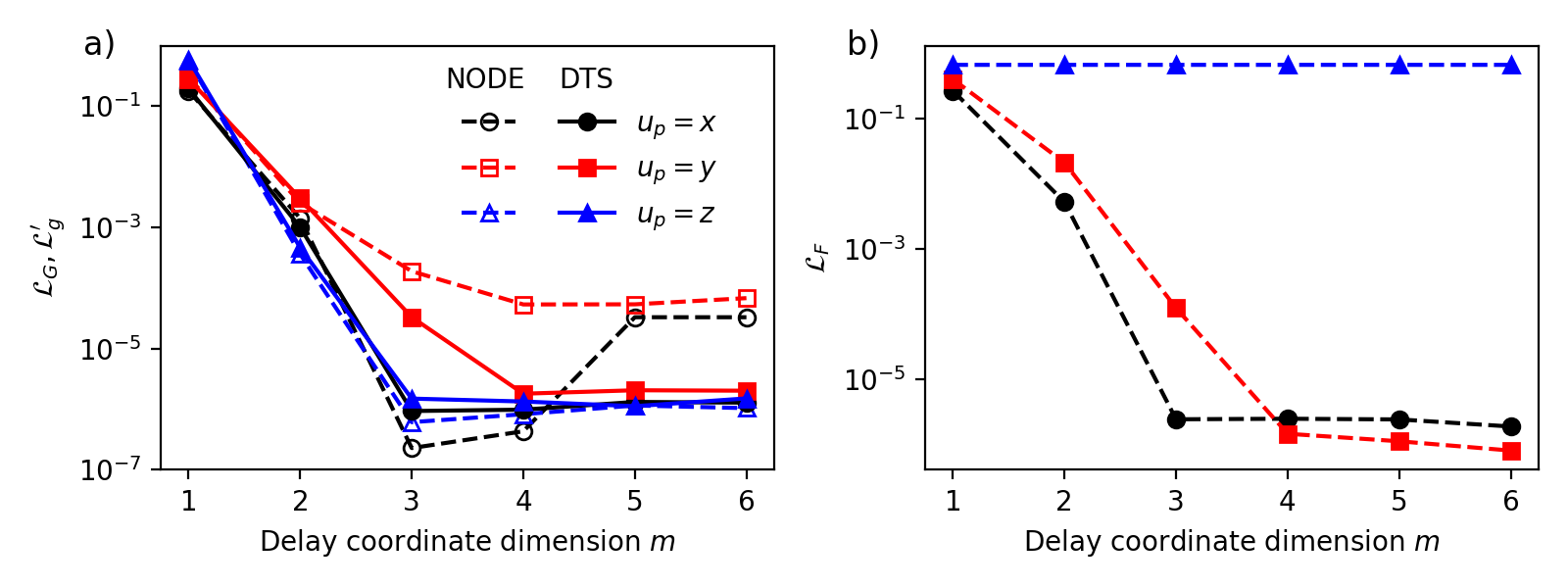}
    \caption{Lorenz attractor test data loss for varying partial observable $u_p$ and embedding dimension $m$ a) Time integration NNs, DTS (solid lines and filled symbols), NODE (dashed lines, open symbols) b) Reconstruction NNs. Delay spacing $\tau = 0.1$ for all models.}
    \label{fig:lorenz_loss}
\end{figure*}

We select $\tau = 0.1$, which is the first minimum of the mutual information (MI). \cite{fraser1986independent}  The embedding dimension determined by FNN is $m = 3$. \cite{kennel1992determining} Time series analysis calculations are performed using the DelayEmbeddings module of the  Julia package DynamicalSystems.jl. \cite{Datseris2018, kramer2021unified} The embedding parameters for different observables $u_p = x,y,z$ were similar. To confirm the choice of $m$ suggested by FNN, we fix the delay spacing and train 5 NNs at a varying embedding dimension $m=1-6$, as shown in Fig. \ref{fig:lorenz_loss}. NN maps for embeddings with delay spacings $\tau = 0.05 - 0.2$ did not qualitatively differ from the presented results. Variance of the MSE is low for both time and reconstruction NNs, although some time stepping models fall onto periodic orbits or fixed points at long times. Therefore, in quantifying the attractor reconstruction (Figs. 3-5) we select the model which best reproduces the attractor joint PDF $P(\tilde{\hat{x}},\tilde{\hat{y}})$ after time integration and reconstruction (Fig. \ref{fig:lorenz_2d}).

The test data mean squared error (MSE) of the DTS map plateaus at an embedding dimension $m = 3$ for an observation $u_p = x$, consistent with FNN and other data-driven approaches. \cite{ouala2020learning} However, we find $m = 4$ is required for $u_p = y$. The need for an additional delay is confirmed by statistical reproduction of the attractor (Figs. 3-5). Observing the $z$-component of the Lorenz appears to provide excellent time prediction, which is unexpected due to the invariance of the Lorenz attractor to the transformation $(x,y,z) \rightarrow (-x,-y,z)$. The low one-step error is misleading, as long-time trajectories generated by the NN fall onto periodic orbits or fixed points. Thus, we consider only the $x$ and $y$ observables in further detail. The MSE of the reconstruction maps are similar to the DTS models, with $u_p = x$ plateauing at $m = 3$ and $u_p = y$ at $m = 4$. The embedding of $u_p = z$ fails to reconstruct the true attractor, as expected and in agreement with previous results using reservoir computing.  \cite{lu2017reservoir}

Comparing DTS and NODE models, we again find an observation $u_p = y$ plateaus at $m=4$. The quantitative value of $L_g ^\prime$ is higher than the DTS model because the NODE is trained to minimize $L_g$. For an observation $u_p = x$ the MSE again decreases up to $m = 3$, although the error increases for $m > 4$. This could be due to error in predicting delay coordinates distance from the current time, or simply due to the introduction of irrelevant information to the embedding. This is consistent with the results of Wang and Guet \cite{wang2021reconstructing}, who found  that applying FNN in an autoencoder reduced an input $m = 6$ embedding of the Lorenz system with observation $u_p = x$ to the leading $m = 3$ delays. Further, they found NODE predictions using an autoencoder performed better in short time tracking than training on the delay coordinate embedding with $m = 6$. However, we find that for the KSE attractor with multidimensional observations FNN does not reliably predict the embedding dimension. Thus, performing this step automatically in an autoencoder without knowledge of the manifold dimension remains challenging.

While the test MSE for the time stepping and reconstruction NNs are low, they capture only the one-step pointwise error. To further quantify the ability of NNs to reconstruct the attractor, we consider ensemble average statistics from long NN trajectories. As expected, both DTS and NODE time integration models with an embedding dimension $m=1$ and $m=2$ go to fixed points and periodic orbits respectively. Therefore, we focus on results for $m \geq 3$.

\begin{figure*}
    \centering
    \includegraphics[width=1.0\linewidth]{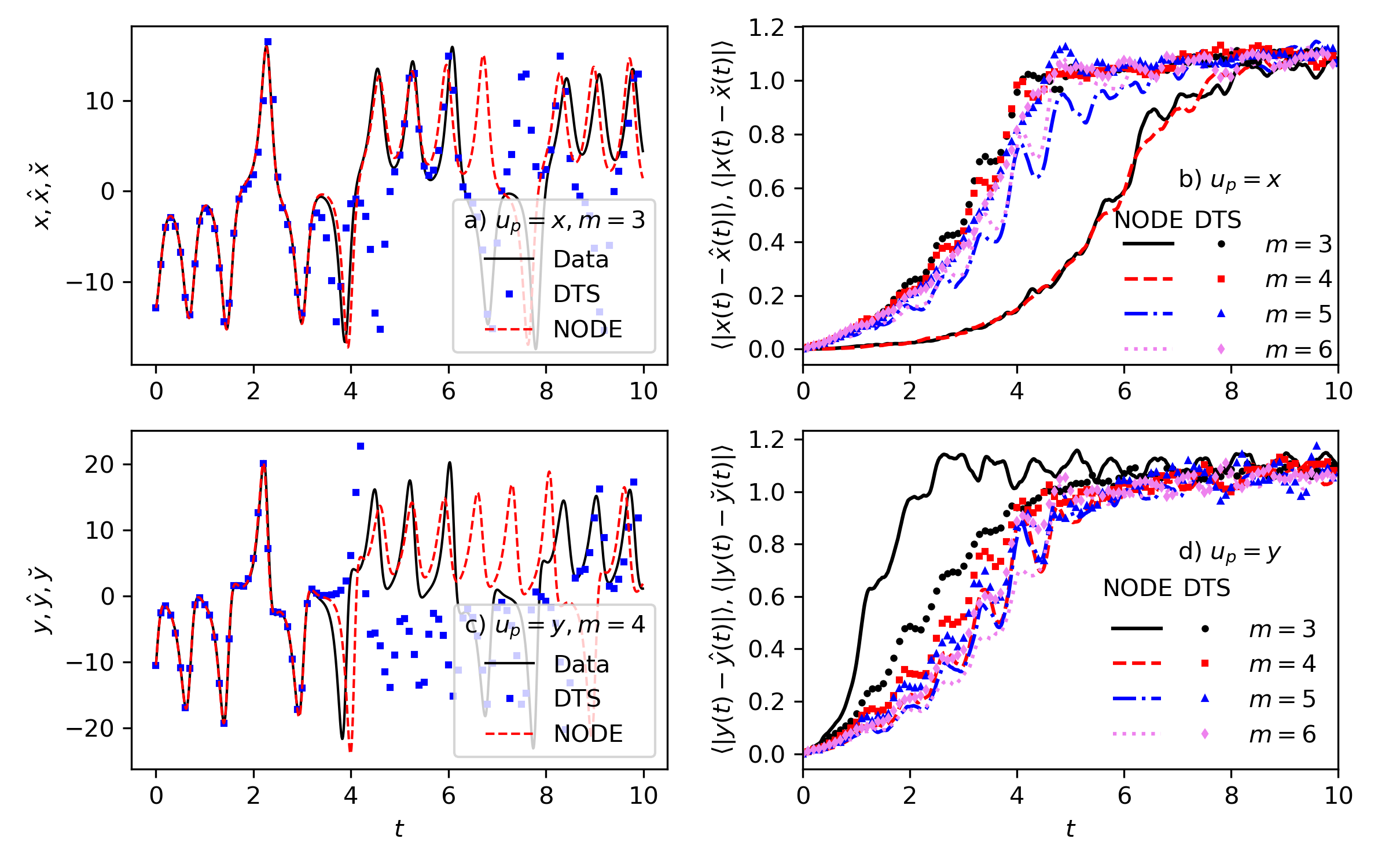}
    \caption{Short term tracking of the Lorenz attractor with different observables (a,b) $u_p = x$ (c,d) $u_p = y$. (left column) Representative trajectories generated from the same initial condition $u_0$. Symbols correspond to predictions from a discrete time step model and dashed lines to a neural ODE model. Both DTS and NODE models use $m = 3, \tau = 0.1$ for $u_p = x$ and $m = 4, \tau = 0.1$ for $u_p = y$.  (right column) Ensemble average error of DTS models (symbols) and NODE models (lines) for increasing embedding dimension. Delay spacing $\tau = 0.1$ for all embeddings.}
    \label{fig:lorenz_tracking}
\end{figure*}

First, we show an example of the short-term forecasting capabilities. We generate 2000 trajectories of the partial observable for 10 time units from different initial conditions using DTS and NODE models. Two representative trajectories are shown in Fig. \ref{fig:lorenz_tracking}a ($u_p = x, m = 3$) and Fig. \ref{fig:lorenz_tracking}c ($u_p = y, m = 4$) where the NNs track for several time units where $\tau_L \approx 1$, comparable to other methods. \cite{ouala2020learning, wang2021reconstructing, gottwald2021combining} We also show the ensemble average tracking error for these two embeddings in Fig. \ref{fig:lorenz_tracking}b,d. Consistent with the test MSE, the tracking error converges for an embedding dimension $m \geq 3$ and observation $u_p=x$. With an observation $u_p=y$, tracking improves slightly with increasing embedding dimension up to $m=6$, although the MSE plateau value $m=4$ already provides good predictive capability and preserves the dynamics in long trajectories (Figs. 4-6).


DTS models for both observables exhibit similar tracking error, but the NODE models for $u_p = x$ are significantly more accurate than for $u_p = y$. We speculate this is due to the sharp gradients in $y$ that occur when the solution jumps from one wing of the attractor to the other, as seen at $t \approx 4.1$ in Fig. \ref{fig:lorenz_tracking}c. Additionally, NODE models observing $u_p = x$ perform worse with $m > 4$, again due to irrelevant information in the embedding. 

Next we quantify the dynamics of a long NN trajectory via the autocorrelation function of the partial observable state
\begin{equation}
	C_p(t) = \frac{\langle u_p(0) \cdot u_p(t) \rangle}{\langle u_p ^2 \rangle}
\end{equation}
\begin{figure*}
    \centering
    \includegraphics[width=1.0\textwidth]{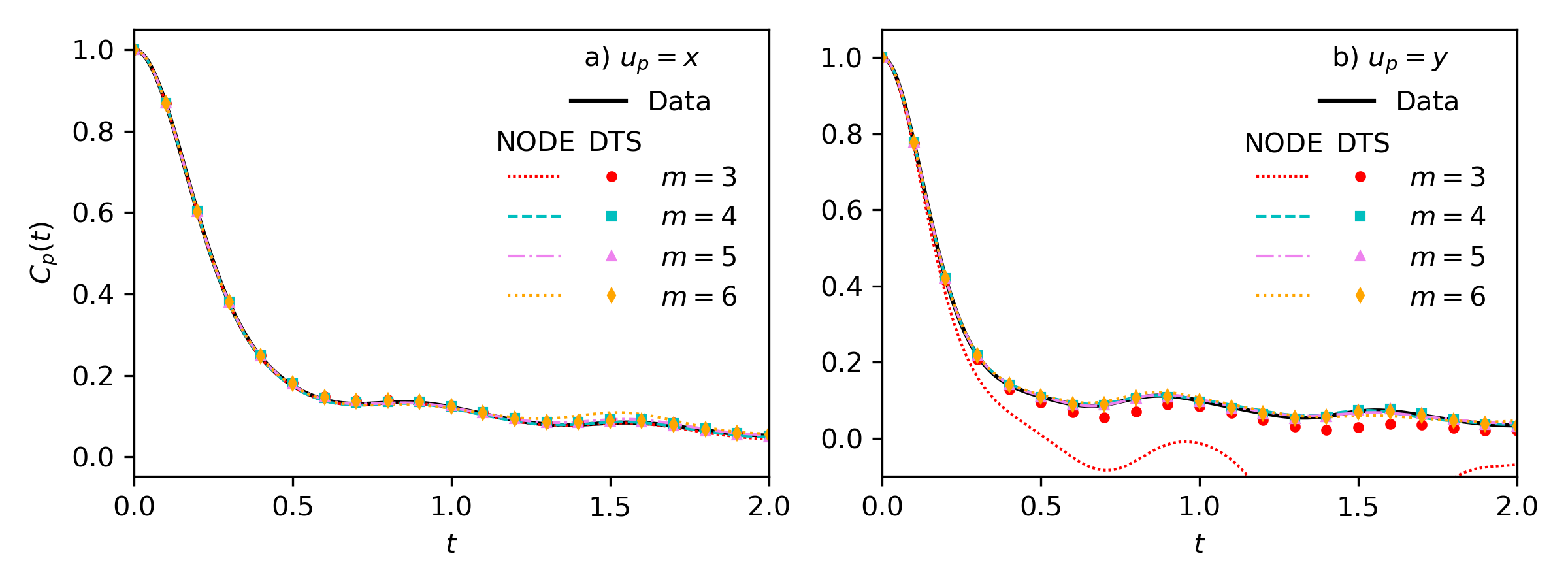}
    \caption{Autocorrelation function of the Lorenz attractor partial observable for a) $u_p = x$ b) $u_p = y$. Solid lines correspond to true data, dashed lines to predictions of NODE models, and symbols to predictions of DTS models. predictions. Delay spacing $\tau = 0.1$. The data and NODE predictions use a sampling interval $\Delta t = 0.01$ and DTS use $\Delta t = \tau$.}
    \label{fig:lorenz_corr}
\end{figure*}
The results shown in Fig. \ref{fig:lorenz_corr} are determined from NN trajectories run for $5 \times 10^4$ time units. For an observation $u_p = x$, the DTS model again reproduces the true data at $m=3$ and achieves similar results with additional delays. For $u_p = y$, the NN model reproduces the data the MSE plateau dimension $m = 4$. The slight improvement up to $m = 6$ found for short-term tracking is not visible in this case. NODE and DTS model predictions agree quantitatively with the data at discrete time step intervals $\tau$, but the NODE models also reproduces the true $C_p(t)$ at arbitrary time scales (sampling $\Delta t = 0.01$ shown here).



We conclude our study of the Lorenz system with the reconstruction of the 3D attractor from the long partial observable trajectories, $\tilde{\hat{u}}$(t), generated by a DTS or NODE model as described in Sec. \ref{sec:recon}. We visualize the reconstruction via the joint PDF $P(x,y)$ in Fig. \ref{fig:lorenz_2d}. The reconstruction results are consistent with other metrics. An observation $u_p = x$ reproduces the joint PDF at $m = 3$ and $u_p = y$ at $m = 4$. We quantify the reconstruction for increasing embedding dimension via the KL divergence in Fig. \ref{fig:lorenz_DKL}
\begin{equation}
	D_{KL}(\tilde{\hat{P}}|P) = \int_{-\infty}^{\infty} \int_{-\infty}^{\infty} P(\tilde{\hat{x}},\tilde{\hat{y}}) \textrm{ln}  \frac{P(\tilde{\hat{x}},\tilde{\hat{y}})}{P(x,y)} 
	\label{eqn:KL_div}
\end{equation}
We assume the contribution to the integral from empty bins is zero as in Ref. \cite{cazais2015beyond}. For reference, we include the KL divergence between two true data sets with different initial conditions (dashed horizontal line). We also show the KL divergence $D_{KL}(\tilde{P} | P)$ for the case of reconstruction of a true partial observable delay coordinate embedding $\tilde{u} = F(u_d; \theta_F)$, which represents a baseline of expected performance for trajectories generated by NN time integration models. All PDFs are generated with the same trajectory length. Using the decoder only, the KL divergence plateaus at the same value as the reconstruction MSE, and the quantitative value is comparable to the divergence of two true data sets. Joint PDFs from the NN integration models have a slight dependence on the number of delays and perform quantitatively worse than only the decoder, as expected. We note the NODEs perform better than the DTS models for $m = 3-6$.

Thus, we have demonstrated that we can learn NN approximations to delay coordinate time integration and reconstruction maps from partial observable data for a low-dimensional ($d_\mathcal{M}  = 3$) chaotic attractor.

\begin{figure*}
    \centering
    \includegraphics[width=1.0\linewidth]{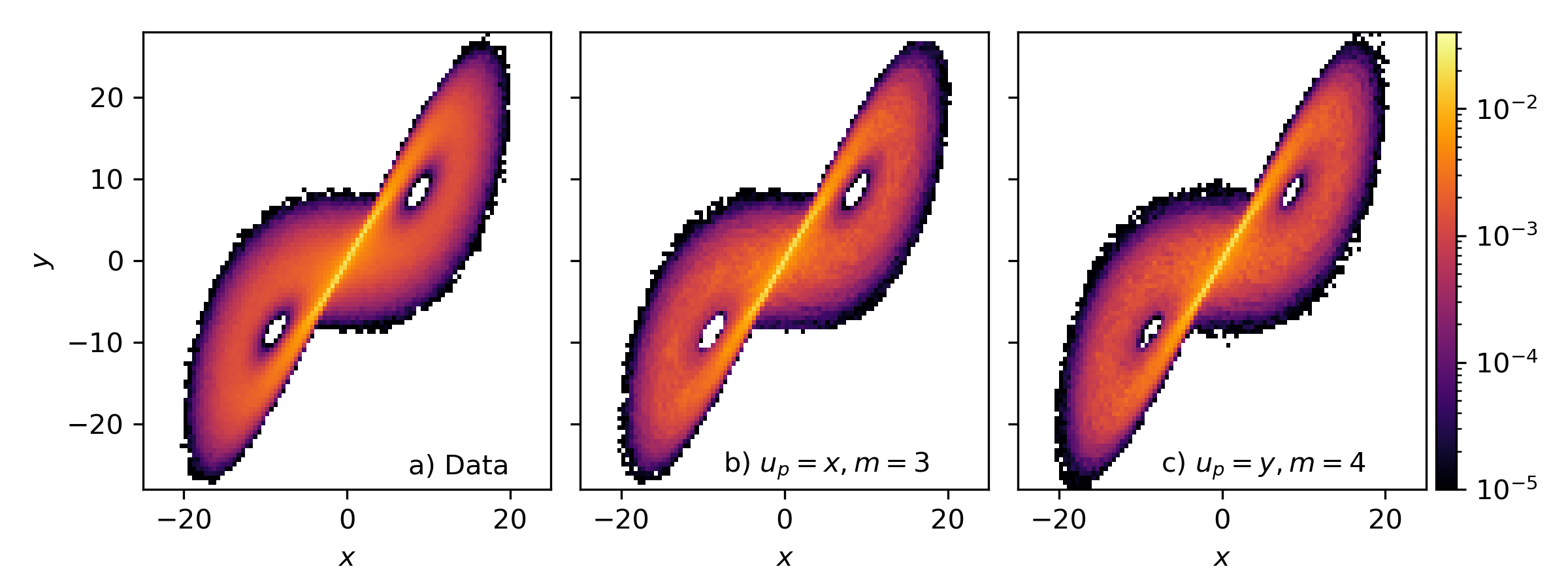}
    \caption{Joint PDF of the Lorenz attractor from a) data $P(x,y)$ and from a discrete time integrated and reconstructed NN trajectory from a delay coordinate initial condition, $P(\tilde{\hat{x}},\tilde{\hat{y}})$  b) $u_p = x, m = 3$ c) $u_p = y, m = 4$. }
    \label{fig:lorenz_2d}
\end{figure*}

\begin{figure}
    \centering
    \includegraphics[width=0.5\textwidth]{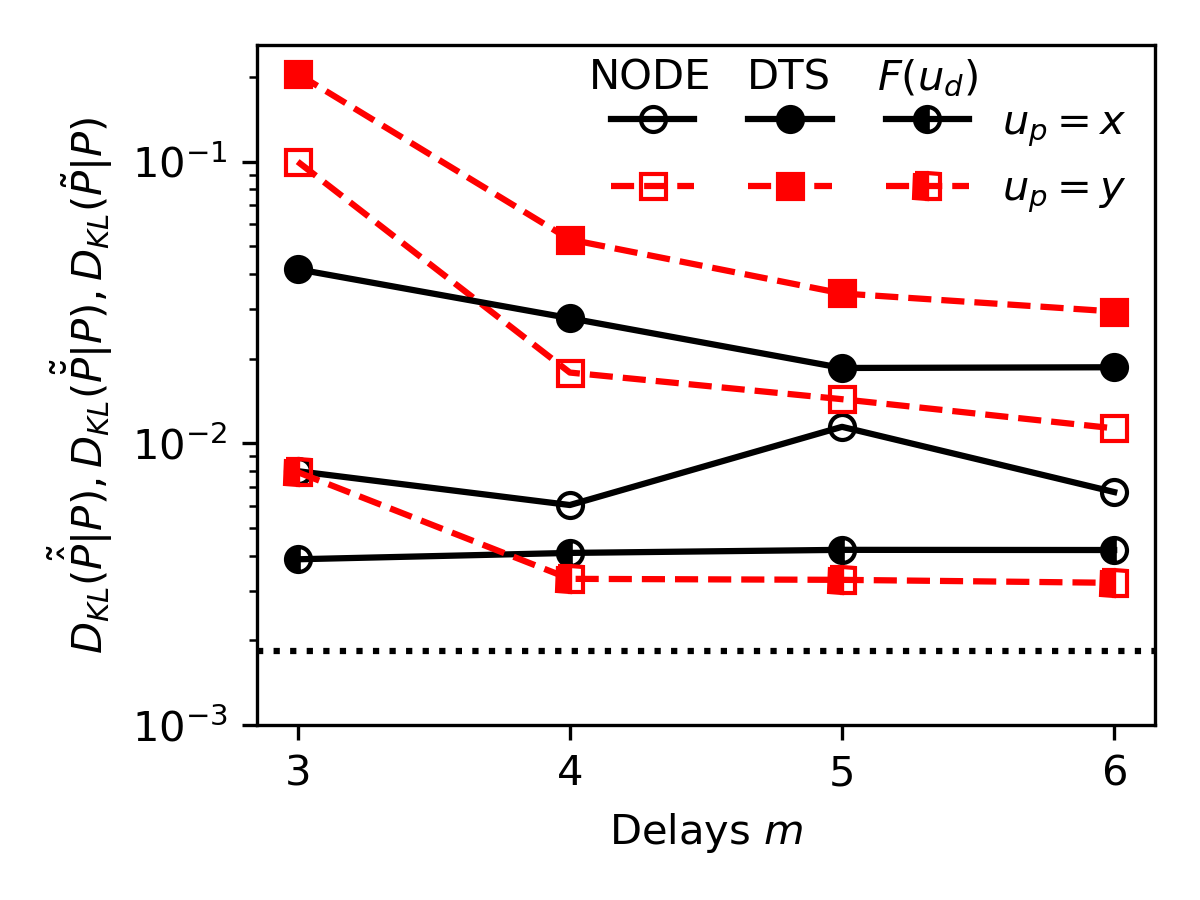}
    \caption{KL divergence of the true and predicted Lorenz joint PDF $P(x,y)$ for increasing number of delays. The horizontal line indicates the divergence between two true data sets with different initial conditions. Open symbols and closed symbols refer to NODE and DTS time integration models respectively. Half-filled symbols refer to reconstruction of a true partial observable trajectory, $F(u_d; \theta_F)$, which excludes error from time integration.}
    \label{fig:lorenz_DKL}
\end{figure}

\subsection{Kuramoto-Sivashinky Equation}

Next we test our method on higher dimensional chaotic attractors with multivariate observations. In particular, we consider the KSE
\begin{equation}
	\partial_t u = -u \partial_x u - \partial_{xx} - \partial_{xxxx} u
\end{equation}
with periodic boundary conditions in the domain $x \in [0,L]$. We consider $L=22,44$ because the manifold dimension for $L = 22$ is known to be $d_\mathcal{M}=8$,\cite{ding2016estimating} and the dynamics become increasingly chaotic with $L$. The manifold dimension for $L=44$ can be approximated by autoencoders \cite{linot2020deep, linot2022data} and the number of physical modes, \cite{yang2009hyperbolicity} which find $d_\mathcal{M} = 18$.  Trajectories were generated using the code from Cvitanovi\'{c} et al. \cite{cvitanovic2005chaos} implementing a Fourier spectral method in space and a fourth-order time integration scheme \cite{kassam2005fourth} on a $d_o = 64$ point grid. We generate a trajectory with $4\times 10^5$ time steps with $\Delta t = 0.25$ with an 80/20 training test/split. DTS and reconstruction NNs are trained by the same procedure as the Lorenz models. NODE models are trained by the same procedure as the Lorenz models, but the number of epochs is increased to 200,000, the learning rate drops every 25,000 epochs, and the number of time steps forecasted during training is $N = 20$. The network depth and width is also increased (Table \ref{table:arch}).

First focusing on $L=22$, we choose evenly spaced grid points as observations in the same manner as Lu et al. \cite{lu2017reservoir} with an observation dimension $d_p = 1,2,4,8$. A diffeomorphism to the state without time delays is expected at $d_p = 16$ due to Whitney's theorem, \cite{guillemin2010differential, sauer1991embedology} so we do not consider $d_p > 8$. Additionally, we will find NNs can forecast and reconstruct the attractor at $d_p = 8$ even without time delays. We need to generate a delay coordinate embedding for each $d_p$. However, generating good embeddings for highly chaotic attractors and multivariate observations is challenging. Several methods have been proposed for multivariate observations, \cite{garcia2005multivariate, nichkawde2013optimal,  kramer2021unified} but in our tests using the code available in DynamicalSystems.jl \cite{Datseris2018, kramer2021unified} they failed to generate an embedding for the KSE with $d_p = 1$. 

Therefore, we estimate a delay spacing via the MI and embedding dimension via FNN for one grid point $d_p =1$, which yield $\tau = 1.5$ and $m = 4$. We use these values as initial guesses and vary both parametrically in training NNs. Currently we consider only uniform embeddings. We find $\tau = 1.0 - 4.0$ to provide the best performance in reconstruction and time stepping. For each $d_p$ we increase the embedding dimension from $m = 1$ to $d_p m = 2 d_\mathcal{M} = 16$, at which dimension we expect to have a diffeomorphism to the state. \cite{deyle2011generalized}

\begin{figure*}
    \centering
    \includegraphics[width=1.0\linewidth]{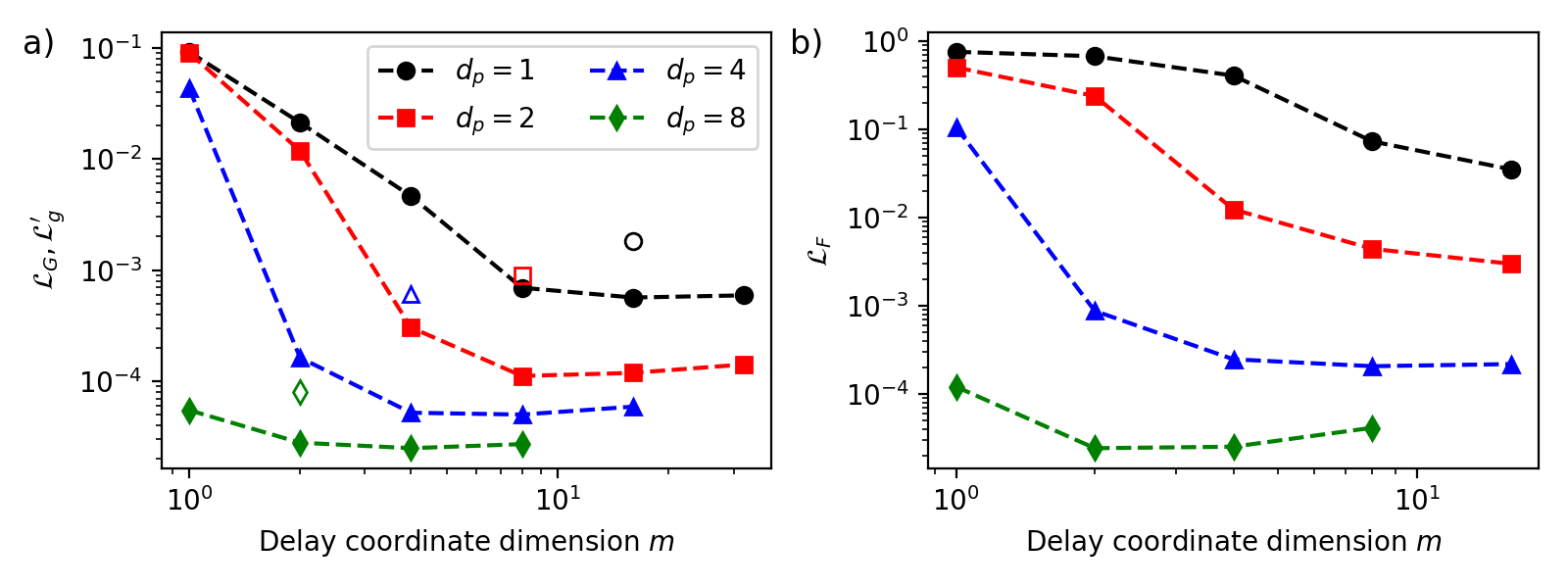}
    \caption{KSE test loss of a) time integration NNs, DTS (filled symbols) and NODE (open symbols) b) reconstructions NNs for increasing observation dimension $d_p$ and delay coordinate dimension $m$. Delay spacing $\tau = 1.5$ for all results. NODE models are only trained for embedding parameters $d_p m = 16 \approx 2 d_\mathcal{M}.$}
    \label{fig:ks_loss}
\end{figure*}

Fig. \ref{fig:ks_loss} shows the test data set loss for the DTS, NODE, and reconstructions models for increasing number of grid points in the observation and number of delays. Here we use the same $\tau = 1.5$ for quantitatively comparing the loss at different $d_p$ because the delay spacing implicitly affects the time step loss through the step size. For later results we will use $\tau = 4.0$ for $d_p = 4,8$ because we find the longer embedding window improves reproduction of attractor statistics. The discrete time NN improves with increasing observation dimension, as expected. Increasing the number of delays $m$ at a fixed $d_p$, we observe a dramatic improvement up to $d_p m = d_\mathcal{M} = 8$ for all observations $d_p$. Providing additional delays, the loss decreases slightly up to $d_p m = 2d_\mathcal{M} = 16$ and then plateaus or slightly increases. 

Trends for the reconstruction map are largely similar, although the decrease in the loss with $d_p$ is more pronounced because the error is now calculated on the full $d_o = 64$-dimensional reconstructed state, rather than the $d_p m$-dimensional partial state. Our results are qualitatively consistent with Lu et al., \cite{lu2017reservoir} who observed a significant improvement in reconstruction from $d_p = 1-4$, and a smaller improvement from $d_p = 8$. Quantitatively, the time-delayed NNs used in this work perform better than reservoir computers (Fig. 8b of Ref. \cite{lu2017reservoir}) which implicitly embed the state history.

Based on the plateau of the DTS and reconstruction model loss at $d_p m = 16$, we train NODE models only for $d_p m = 16$. As noted in Sec. \ref{sec:NODEs}, we calculate the DTS loss term using the NODEs $L_g'$ for comparison of the two models. The NODE error also decreases as the dimension of the observation $d_p$ increases. The quantitative value of $L_g ^\prime$ is larger than $L_G$, again because NODEs are trained to minimize a different loss. In comparing tracking and attractor reconstruction below we find the NODEs perform well.

\begin{figure}
    \centering
    \includegraphics[width=0.5\textwidth]{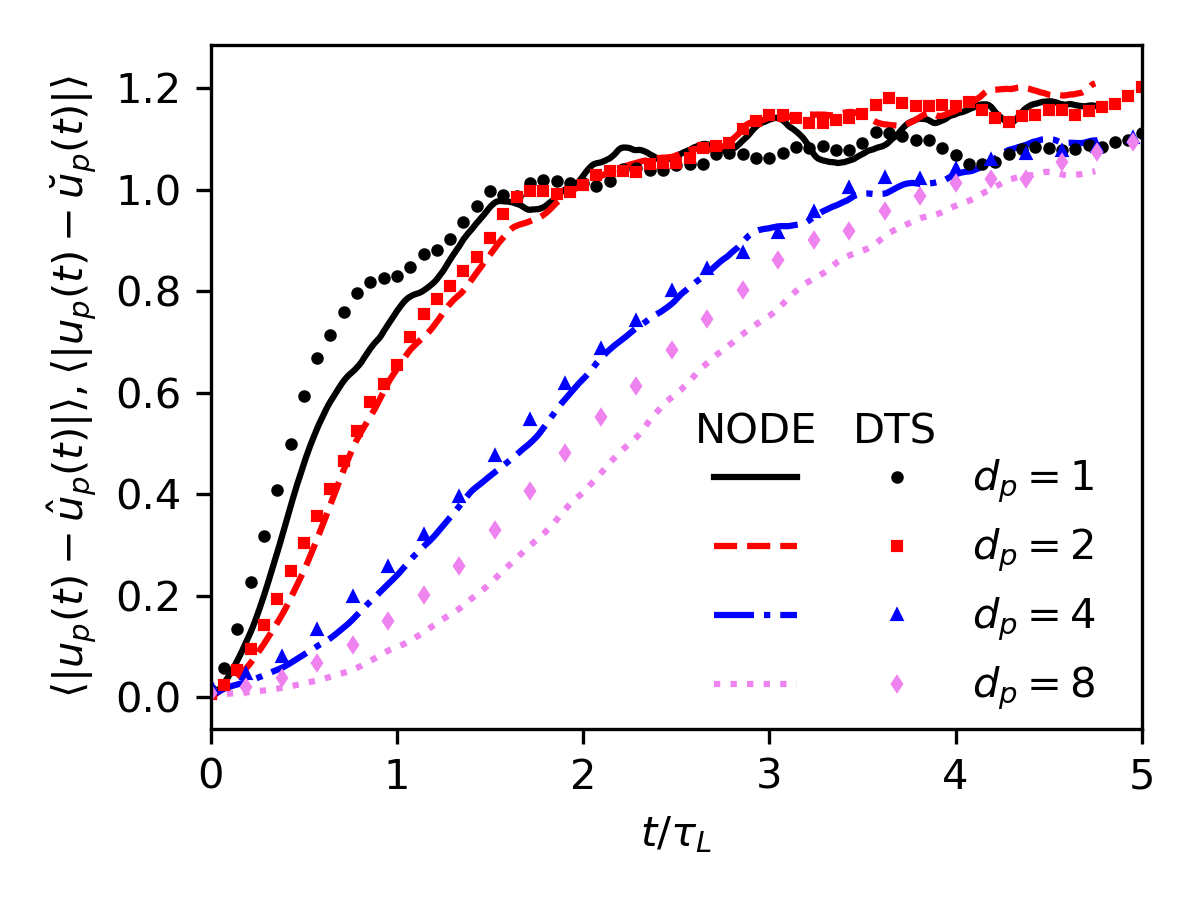}
    \caption{Short term ensemble average tracking of the KSE partial observable for NODE models (lines) and DTS models (points) with increasing observation dimension $d_p$. Delay spacing $\tau = 1.5$ for $d_p = 1,2$ and $\tau = 4.0$ for $d_p = 4,8$. The product of number of time delays $m = 16,8,4,2$ and $d_p = 1,2,4,8$ is constant, $d_p m = 16 \approx 2 d_\mathcal{M}.$}
    \label{fig:ks_ens_track}
\end{figure}

We quantify the ensemble average tracking error of the partial state for increasing observation dimension $d_p$ in Fig. \ref{fig:ks_ens_track}. Here we show only one embedding model for each $d_p$ corresponding to $d_p m = 16$. As noted above, we find that larger delay spacings perform better for larger observation dimensions due to the increased delay window, so in these results we use $\tau = 4.0$ for $d_p = 4,8$. We see that for a sparse observation of the state space, both DTS and NODE models diverge from the true solution relatively quickly compared to the Lyapunov time $\tau_L \approx 21$. \cite{ding2016estimating} Tracking improves dramatically from $d_p = 2$ to $d_p = 4$ and slightly more for $d_p = 8$. For $d_p = 4,8$ there is a modest dependence on $\tau$, but for $d_p = 1,2$ predictions separate from the true solution quickly regardless of the choice of $\tau$. We generally find NODE models to perform better than or equivalent to DTS models in short term tracking. This could be related to the NODEs being trained with a data spacing $\Delta t = 0.25$ as compared to $\tau = 1.5-4.0$ for the DTS models,  although discrete time steppers trained on ROMs of the KSE full state have shown prediction degradation does not occur until a data spacing $0.4 \tau_L \approx 8.4$. \cite{linot2022data}

\begin{figure*}
    \centering
    \includegraphics[width=\textwidth]{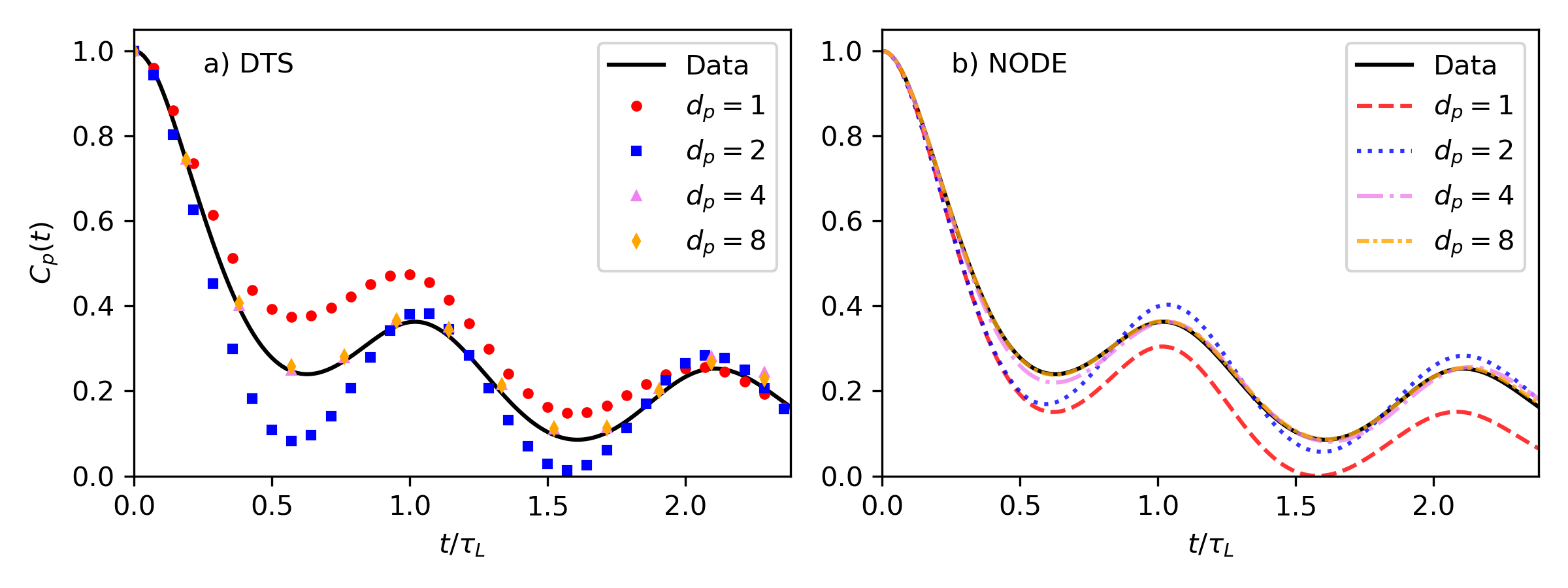}
    \caption{Autocorrelation function of the KSE partial observable for increasing observation dimension $d_p$. Results from a) DTS models b) NODE models. Embedding parameters are the same as Fig. \ref{fig:ks_ens_track}.}
    \label{fig:ks_corr}
\end{figure*}

Next we investigate the long time dynamics of the NN discrete time maps by the autocorrelation in Fig. \ref{fig:ks_corr}, again showing only one embedding for each $d_p$. The data is generated from trajectories run for $t \approx 1 \times 10^5$ time units or $t/\tau_L \approx 5000$. Both DTS and NODE model predictions for $d_p = 4,8$ reproduce the data up to 2-3 Lyapunov times. These observables also match the true autocorrelation up to at least $\tau_L$ for lower embedding dimensions $d_p m = d_\mathcal{M} = 8$. Shorter or longer delay windows from $\tau = 1.0-6.0$ also agree up to $\tau_L$. However, model predictions using $d_p = 1,2$ do not reproduce the data for any combination of hyperparameters and embedding parameters we tested. This suggests there is a number of observables at which learning the time map becomes significantly easier, which is consistent with the improvement from $d_p = 2$ to $d_p = 4$ seen in Ref. \cite{lu2017reservoir}. This could be related to the number of determining nodes as predicted by inertial manifold theory \cite{foias1991determining} and infinite dimensional versions of Takens theorem \cite{foias1995determining, kukavica2004distinguishing}, although these works predict a significantly lower dimensional observation of $d_p = 4, m = 1$ or $d_p = 1, m = 4$ fully describe the KS attractor. 

\begin{figure}
    \centering
    \includegraphics[width=0.5\textwidth]{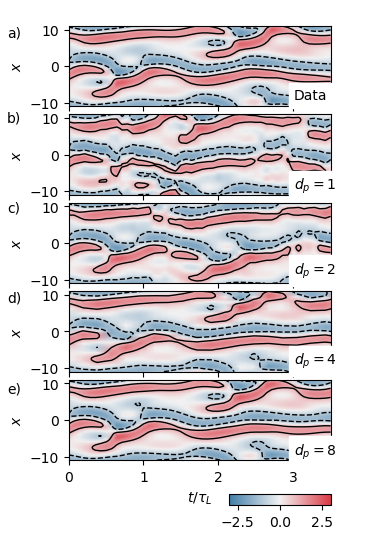}
    \caption{Visualization of KSE reconstruction after time integration using DTS models. Color contour trajectories $u(x,t)$ with solid lines at $u=1$ and dashed lines at $u=-1$ a) Data b) $d_p = 1, m = 16, \tau = 1.5$ c) $d_p = 2, m = 8, \tau = 1.5$ d) $d_p = 4, m = 4, \tau = 4.0$ e) $d_p = 8, m = 2, \tau = 4.0$.}
    \label{fig:ks_recon_track}
\end{figure}

The success of NNs with $d_p = 4,8$ could simply be because the diffeomorphism is easier to learn, as compared to $d_p = 1,2$ which are theoretically diffeomorphic to the attractor at $m = 16,8$, but are found to perform significantly worse in practice. Both deviate from the true correlation function after $0.25 \tau_L$, although the dynamics remain chaotic. The poor long time performance is despite the fact that the one step loss for $d_p = 2, m = 8$ is quantitatively comparable to $d_p = 4, m = 4$. This suggests the predicted trajectory initially stays on the true attractor for some short duration, which we visualize below after attractor reconstruction (Fig. \ref{fig:ks_recon_track}b).

We comment that partial observation embeddings $d_p = 1,2,4$ with few delays perform significantly worse than the results shown here both in short term tracking and long time dynamics. Generally for $d_p m < d_\mathcal{M}$, the predicted trajectory quickly goes to a fixed point or periodic orbit. While $d_p = 1,2$ do not quantitatively reproduce the attractor, there is a clear improvement from the delay coordinate embedding. 

%

Finally, we investigate the quality of the reconstruction of the true KS attractor simulated on a grid from our delay coordinate embeddings. As with the Lorenz attractor, we first consider an individual trajectory to visualize the effect of increasing observation dimension. In Fig. \ref{fig:ks_recon_track}, the same initial condition is used in each panel. The true data (Fig. \ref{fig:ks_recon_track}a) is generated from numerical integration of the KSE. The panels below are generated by first filtering the initial condition to the appropriate number of equally spaced grid points $d_p$ and embedding these observations with time delays of the partial state. The trajectories forward for $t \approx 70$ time units as detailed in \ref{sec:timestep} and reconstructed to the full state as in \ref{sec:recon}. A DTS model is used here, but the NODE model predictions are visually similar.

For the trajectory shown, the NN time integrated and reconstructed trajectory for $d_p = 8, m = 2$ shows excellent agreement with the true solution up to 70 time units. The trajectory generated from $d_p = 4, m = 4$ also performs well, although visible differences emerge by $2 \tau_L \approx 40$. Even a sparse observation $d_p = 2, m = 8$ generates reasonable tracking and reconstruction up to $\tau_L$, although at intermediate times it becomes clear the predicted solution leaves the true attractor. The single grid point $d_p = 1$ time prediction separates from the true solution by $0.5 \tau_L$, and error in the reconstruction are visually apparent.


\begin{figure*}
    \centering
    \includegraphics[width=1.0\linewidth]{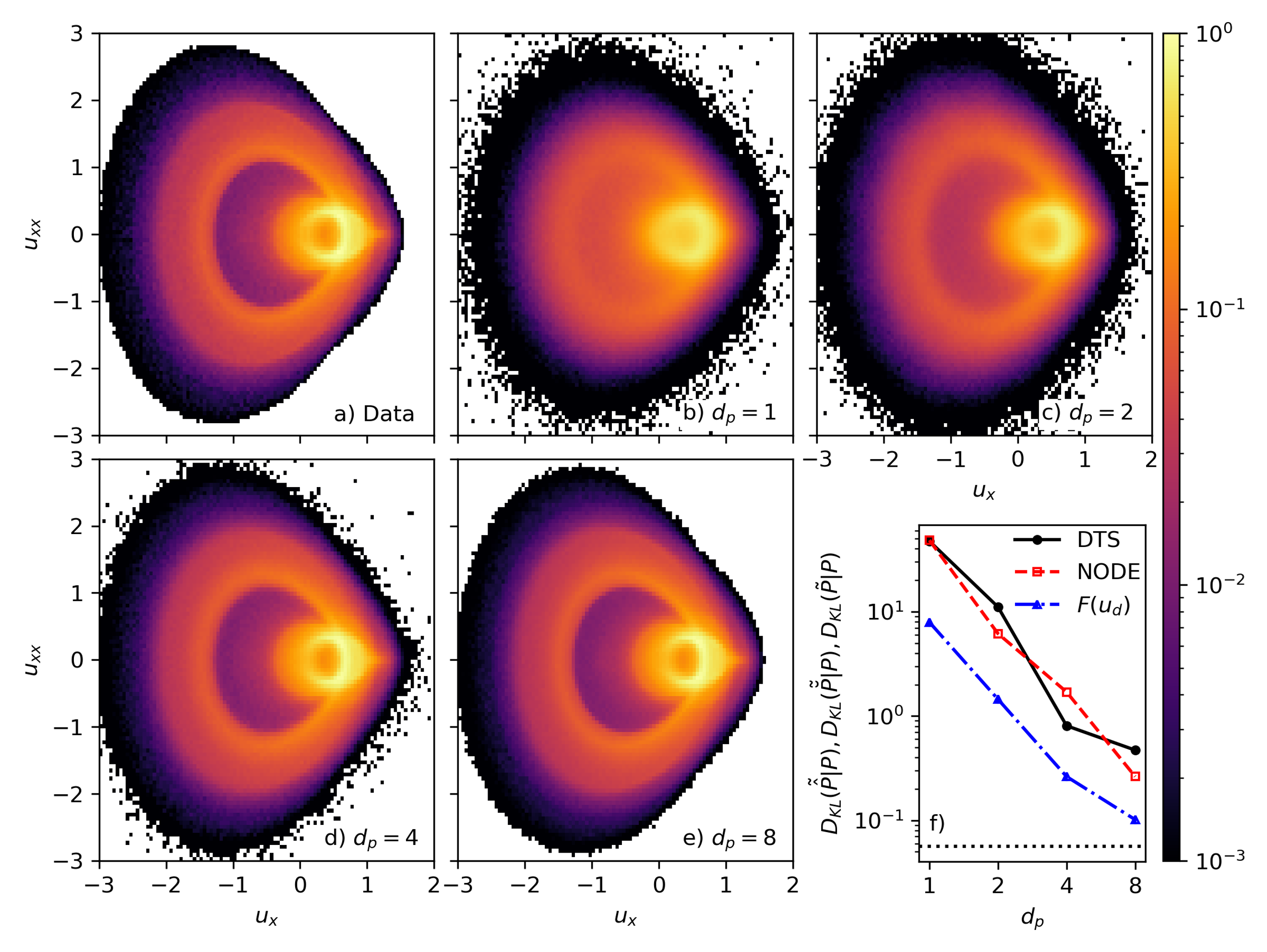}
    \caption{Joint PDFs of the KSE for $L = 22$ generated from a) data and discrete time integrated and reconstructed NN trajectories, $\tilde{\hat{u}} = F(G(u))$, with observation dimension and observation dimension b) $d_p = 1$ c) $d_p = 2$ d) $d_p = 4$ e) $d_p = 8$. Embedding parameters are the same as previous figures. f) KL divergence of the true attractor PDF and NN reconstructed PDFs. Filled symbols refer to DTS forecasted data, open symbols refer to NODE forecasted data, and half-filled symbols to reconstruction of a true partial observable trajectory without time integration. The horizontal dotted line indicates the KL divergence of two PDFs from numerical simulations with different initial conditions.}
    \label{fig:ks_pdf_DKL}
\end{figure*}

Next we visualize the accuracy of reconstruction of NN time integrated trajectories used to generate the autocorrelation function in Fig. \ref{fig:ks_corr}.  We consider the joint PDF of the first and second spatial derivatives, $P(u_{x},u_{xx})$ (Fig. \ref{fig:ks_pdf_DKL}), which provides a detailed view of the attractor. \cite{linot2020deep, linot2022data} Joint PDFs generated by a DTS model are shown and NODE results are similar. Attractor reconstruction again improves with the observation dimension, with $d_p = 8$ showing excellent agreement. With $d_p = 4$, there are small and rare excursions off the true attractor, but the finer details are retained. At lower dimensions $d_p = 1,2$, the predicted values do not capture the high density regions of the attractor accurately, in agreement with the other metrics. We quantify the difference between the predicted and true attractors with the KL divergence $D_{KL}(\tilde{\hat{P}} | P)$, defined similarly to the Lorenz case (Eqn. \ref{eqn:KL_div}). The predictions improve by over an order of magntiude from $d_p = 2$ to $d_p = 4$, and then by a factor of two for $d_p = 8$. The KL divergence of the true data and model predictions are comparable for both DTS and NODE time integration. We include a comparison to the KL divergence $D_{KL}(\tilde{P} | P)$ of a joint PDF generated by reconstruction of a true partial trajectory without time integration, $F(u_d; \theta_F)$. As expected it is closer to the true data due to the lack of time integration error. The dashed horizontal line indicates the KL divergence between two true solutions with different initial conditions, which is quantitatively comparable to the $d_p = 8$ prediction.

\begin{figure*}
    \centering
    \includegraphics[width=1.0\linewidth]{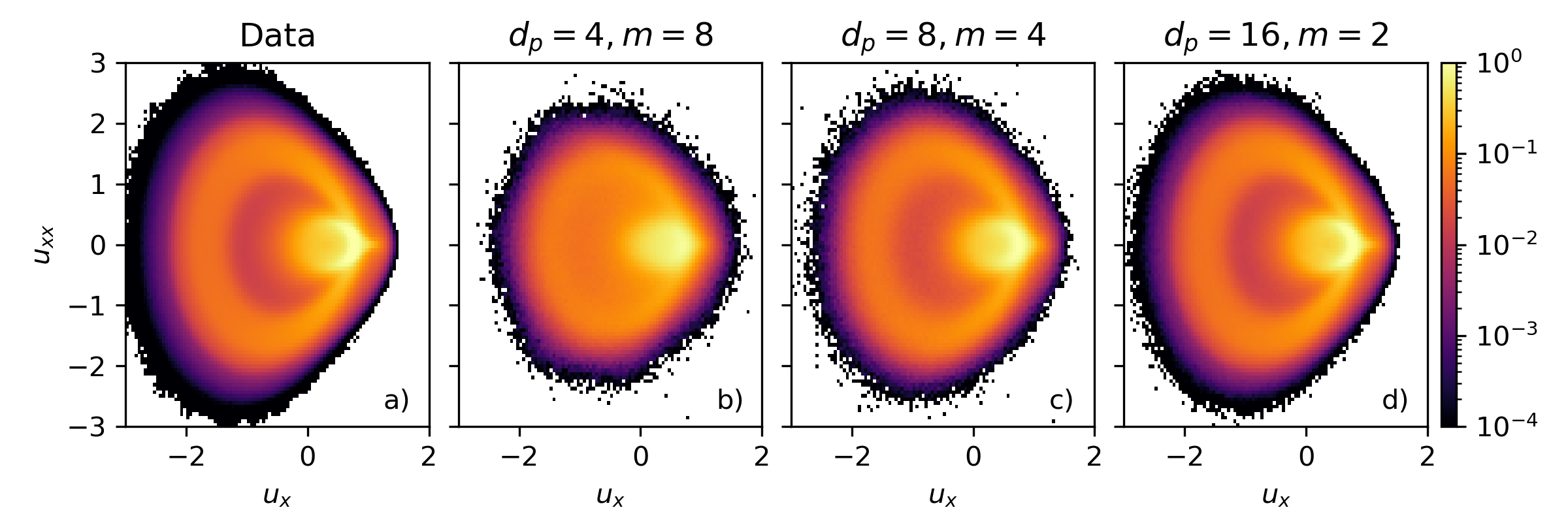}
    \caption{Joint PDFs of the KSE for $L = 44$ as determined from a) Data b) $d_p = 4, m = 8, \tau = 1.25$ c) $d_p = 8, m = 4, \tau = 2.5$ d) $d_p = 16, m = 2, \tau = 2.5$. Time integration is performed using a DTS model b-d.}
    \label{fig:L44pdf}
\end{figure*}
\begin{figure}
    \centering
    \includegraphics[width=0.5\textwidth]{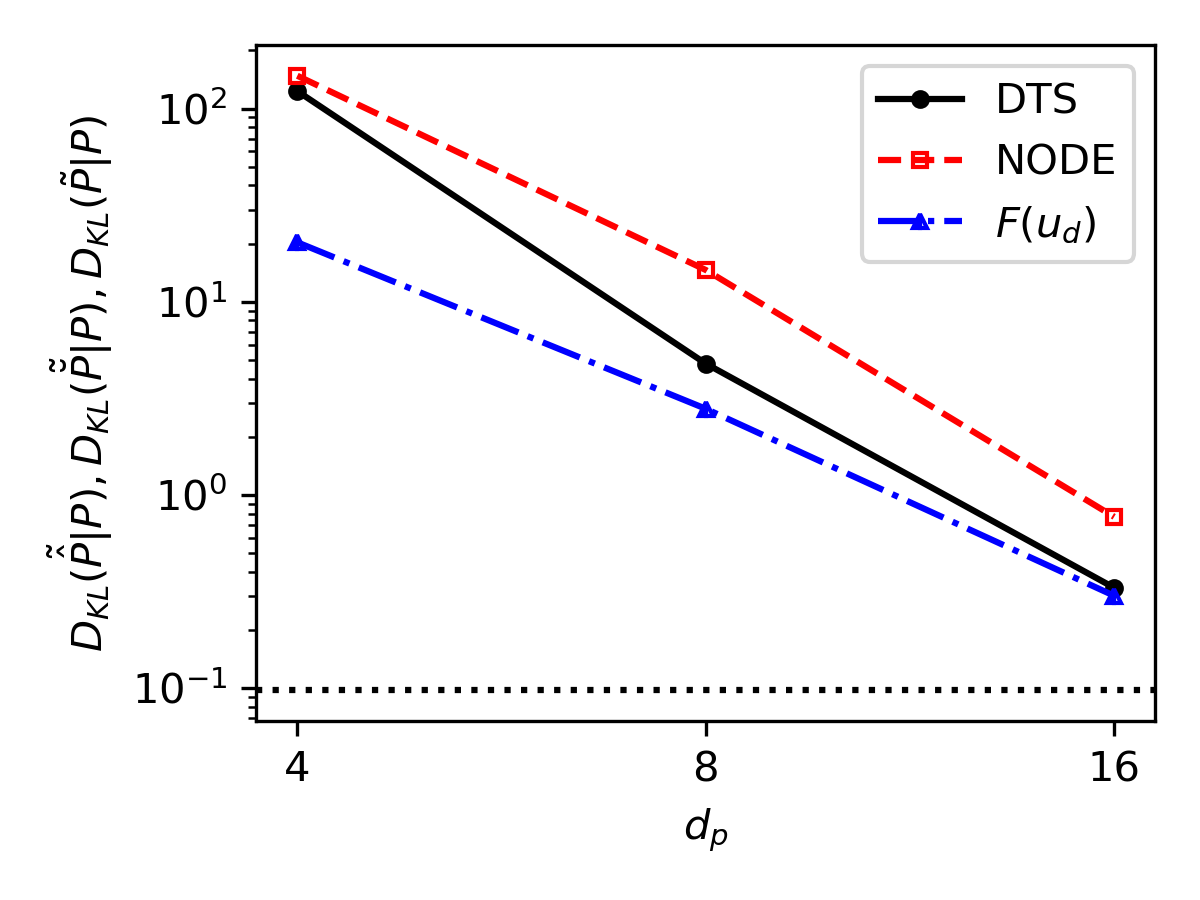}
    \caption{KL divergence of the true and predicted KS $L=44$ joint PDF $P(u_x,u_{xx})$ for increasing observation dimension $d_p$. All embedding dimensions are $d_p m = 32$. The horizontal line indicates the divergence between two true data sets with different initial conditions. Open symbols and closed symbols refer to NODE and DTS time integration models respectively. Half-filled symbols refer reconstruction of a true partial observable trajectory, $F(u_d; \theta_F)$, which excludes error from time integration.}
    \label{fig:L44_dkl}
\end{figure}

To further demonstrate the scaling of our approach to higher dimensional attractors, we consider data from the KSE with $L = 44$, which lies on a manifold of dimension $d_\mathcal{M} = 18$. The numerical simulation details, amount of training data, and NN training procedure are the same as for $L = 22$. The network width is increased to provide additional capacity for modeling the higher dimensional attractor. Here we show only the joint PDFs $P(u_{x},u_{xx})$ in Fig. \ref{fig:L44pdf} and the associated KL divergence for DTS, NODE, and reconstruction models in Fig. \ref{fig:L44_dkl}.

We again find that as the observation dimension $d_p = 16$ approaches the attractor dimension, the NNs are successful even without delays, $m=1$, although an additional delay $m = 2$ provides quantitative improvement. At $d_p = 8 \approx 1/2 d_\mathcal{M}$, the NN predictions stay on the attractor at an embedding dimension $m = 4$, although the predictions are quantitatively worse than the comparable case at $L =22, d_p = 4 \approx 1/2 d_\mathcal{M}$. At lower dimensions $d_p = 4 \approx 1/4 d_\mathcal{M}$, the NNs again fail to provide accurate predictions for any delay embedding dimension. Thus there may be practical limitations in learning global delay coordinate maps for sparse measurements on high dimensional attractors. This could be alleviated by a multiple charts and atlases approach, in which the attractor is clustered into regions which may be locally lower dimensional and thus easier to approximate. \cite{floryan2021charts} DTS models are qualitatively comparable to NODE models but more quantitatively accurate in reproducing attractor statistics (Fig. \ref{fig:L44_dkl}).

\section{\label{sec:Conclusions}Conclusions}

We have presented a method for forecasting and reconstructing chaotic attractors from partial observable data. We use deep neural networks to learn functions that approximate the diffeomorphic mapping from a delay coordinate embedding to the true attractor. We have verified the approach on two common model systems: the 63 Lorenz system and the KSE. The low-dimensional Lorenz model can be accurately predicted in time and reconstructed to the true state from a scalar observation. The KSE, however, requires multivariate observations to fully reproduce the attractor. We have tested our method on short time tracking and long time dynamics. 

The method has similarities to other data-driven approaches which reconstruct latent attractors and learn discrete time maps or continuous time flows. However, we have demonstrated the capacity of DNNs to reproduce high-dimensional attractors via the KSE at $L=22,44$. Currently it has only been applied to embeddings with uniform time delays, but it can also be applied to non-uniform embeddings. In this case neural ODEs would be advantageous because the discrete time map would require interpolation or a restrictive choice of time step to update the time delays. A limitation of uniform embeddings is that it is difficult to learn the delay coordinate maps for sparse observations with many delays, as evidenced by the relatively poor performance for KSE with $d_p < 1/2 d_\mathcal{M}$. In this case, information at times intermediate to the uniform delays could improve the quality delay coordinate phase space, although our attempts on the KSE using several existing methods \cite{garcia2005multivariate, nichkawde2013optimal,  kramer2021unified} did not converge.

Our method is relevant to applications requiring forecasting of partial observations from experimental data. Reconstruction would require data from a numerical simulation to perform the supervised learning process. We are particularly interested in applying our approach to control of turbulent flows by reinforcement learning. \cite{brunton2020machine, ren2021applying, zeng2021symmetry, zeng2022data} These control policies often use highly resolved state space data, but in experiments data sampling is often poor away from the wall. Delay coordinate embeddings could provide an alternative state space for control, as has been demonstrated for the $L=22$ KSE with sensors at 8 grid points \cite{bucci2019control} and nonlinear underactuated systems. \cite{knudsen2022model}

\begin{acknowledgements}
This work was supported by AFOSR  FA9550-18-1-0174 and ONR N00014-18-1-2865 (Vannevar Bush Faculty Fellowship).
\end{acknowledgements}

\bibliography{main}

\end{document}